\documentclass[lettersize,journal]{IEEEtran}
\usepackage{subcaption}
\usepackage{amsmath,amsfonts}
\usepackage{algorithmic}
\usepackage{algorithm}
\usepackage{array}
\usepackage{textcomp}
\usepackage{stfloats}
\usepackage{url}
\usepackage{verbatim}
\usepackage{graphicx}
\usepackage{cite}
\begin{document}

\title{BCG-Former: Toward Pareto-Efficient Hyperspectral Image Classification via Band-Contextual Gating}

\author{Gaurav~Sharma,~Eungjoo~Lee,~\IEEEmembership{Member,~IEEE}
\thanks{G. Sharma is with the College of Information Science, University of Arizona, Tucson, AZ, USA.}
\thanks{E. Lee is with the Department of Electrical and Computer Engineering, University of Arizona, Tucson, AZ, USA (corresponding author: eungjoolee@arizona.edu).}}





\maketitle

\begin{abstract}
Hyperspectral image (HSI) classification systems are increasingly deployed on platforms with strict computational budgets, such as UAVs and small spaceborne sensors. In these settings, accuracy alone is not enough; the model must also run within tight latency and memory constraints. Most recent HSI classifiers, however, focus on accuracy and pay relatively little attention to these constraints. We propose \textbf{BCG-Former}, a lightweight CNN-Transformer hybrid that targets this trade-off. The model introduces three innovations: (1) Band-Contextual Gating (BCG) for adaptive spectral recalibration using local inter-band context and learnable temperature sharpening, (2) a spectral summary token that bridges spectral and spatial features, and (3) single-pass Band-RoPE combined with linear attention for efficient joint representation learning.
Evaluated on classical airborne (Pavia University, Salinas, Indian Pines, Houston 2013/2018) and UAV-borne benchmark datasets (WHU-Hi-LongKou, HongHu, and HanChuan), BCG-Former achieves overall accuracy ranging from 91.51\% on Houston 2018 to 99.49\% on Houston 2013, while maintaining sub-millisecond inference latency (0.91--0.95ms) and using only 0.10--0.23M parameters. Across all eight benchmarks, BCG-Former consistently resides on or near the Pareto frontier of accuracy versus latency, outperforming or matching recent CNN-, Transformer-, and Mamba-based methods at a fraction of their computational cost. Ablation studies confirm that all three components are complementary, with BCG providing the largest individual contribution. These results establish BCG-Former as a strong accuracy-efficiency Pareto candidate for real-time and large-scale remote sensing applications.
\end{abstract}


\begin{figure*}[!t]
    \centering
    \includegraphics[width=0.98\textwidth]{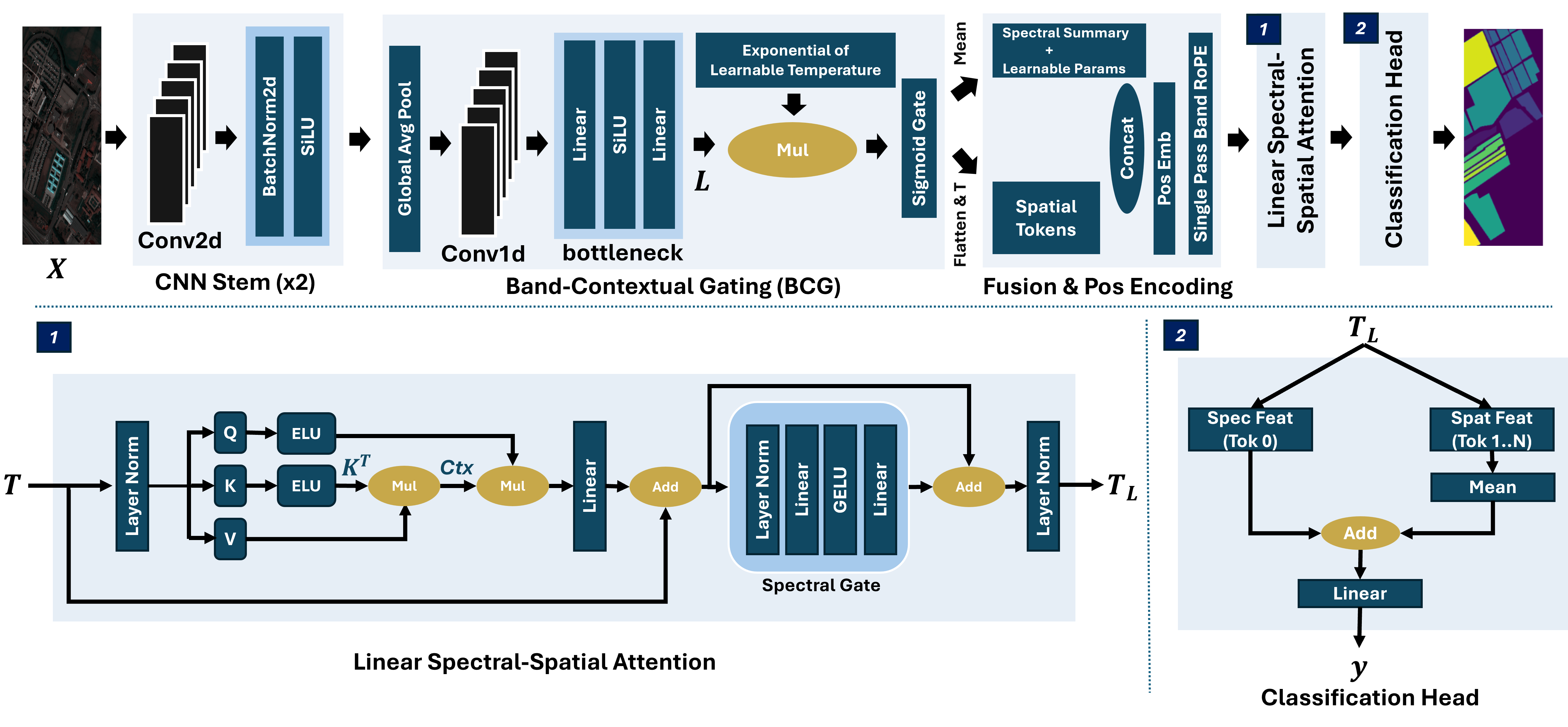}
   \caption{Overview of the proposed BCG-Former framework for efficient hyperspectral image classification, highlighting the Band-Contextual Gating module, spectral summary token, and single-pass Band-RoPE with linear attention.}
    \label{fig:architecture}
\end{figure*}

\section{Introduction}

Hyperspectral image (HSI) classification is a cornerstone of modern remote sensing, leveraging rich spectral signatures from hundreds of contiguous narrow bands to enable fine-grained material discrimination beyond the capabilities of multispectral sensors \cite{richards2013remote, zhang2015hyperspectral}. This makes it essential for applications such as land-cover mapping, precision agriculture, urban analysis, and environmental monitoring. However, effectively modeling the 3D HSI datacube remains challenging, as it requires the joint extraction of local spatial textures and long-range spectral dependencies \cite{yun2023spectr}. These challenges are further compounded by the high dimensionality of hyperspectral data, which introduces strong spectral redundancy and highly correlated feature spaces, aggravating the curse of dimensionality. Additionally, the limited availability of labeled samples in standard benchmarks increases the risk of overfitting and necessitates compact yet expressive models \cite{liu2020dimensionality, demir2014active}.

Early deep learning methods primarily relied on convolutional neural networks (CNNs) such as MobileNet \cite{mobilenet} and HybridSN~\cite{hybridsn}. While computationally efficient, CNNs are inherently limited by their fixed receptive fields, struggling to capture long-range spectral dependencies. Vision Transformers (ViTs)-based models like SSFTT~\cite{ssftt} address this limitation through global self-attention, but their quadratic complexity $O(N^2)$ becomes prohibitive as sequence lengths grow. Recent hybrid approaches, such as HiT \cite{hit} and Swin-HSI \cite{swinhsi}, attempt to balance efficiency and performance, yet they often discard important spectral ordering due to fixed patch tokenization. More recently, state-space models such as SpectralMamba \cite{spectralmamba} have achieved linear complexity for spectral modeling, but they provide limited spatial reasoning capability.

Despite these advances, a significant gap remains where, to the best of our knowledge, no existing method simultaneously achieves high classification accuracy, sub-millisecond to low-millisecond inference latency, and with only 0.10-0.23\,M parameters consistently across multiple diverse HSI benchmarks. Critically, prior works optimize accuracy in isolation, neglecting the accuracy-latency Pareto frontier that determines real-world deployability. To address this gap, we propose \textbf{BCG-Former} (Fig.~\ref{fig:architecture}), a lightweight CNN-Transformer hybrid for HSI classification. The model is built around hardware-friendly operations and HSI-specific inductive biases, rather than scaled depth or parameter count. We evaluate BCG-Former on eight benchmark hyperspectral datasets covering both airborne (Pavia University~\cite{paviauniversity}, Salinas~\cite{salinas}, Indian Pines~\cite{indian_pines_1992}, Houston 2013~\cite{houston}/2018~\cite{houston18}) and UAV-borne (WHU-Hi-LongKou~\cite{whu_hi_datasets}, HongHu~\cite{whu_hi_datasets}, and HanChuan~\cite{whu_hi_datasets}) sensing scenarios. The main contributions of this work are:
\begin{itemize}
\item \textbf{Band-Contextual Gating (BCG)}, an HSI-specific spectral recalibration module that adaptively re-weights channels using local inter-band context and learnable temperature sharpening, providing consistent accuracy improvements with negligible computational overhead.

\item \textbf{A spectral summary token} that pools the BCG-gated features into a single global descriptor and is concatenated with the spatial tokens. This gives the transformer direct access to aggregated spectral information when it computes attention.

\item \textbf{Single-pass Band-RoPE with ELU-kernel linear attention.} Rotary positional encoding is adapted to the ordered spectral-band token sequence and applied once before the Transformer blocks, enabling efficient relative inter-band positional modeling. Combined with ELU-kernel linear attention, this reduces complexity from $\mathcal{O}(N^{2}D)$ to $\mathcal{O}(ND^{2})$.

\item \textbf{A strong accuracy-efficiency trade-off}, validated across five airborne datasets (Pavia University, Salinas, Indian Pines, Houston 2013/2018) and three UAV-borne datasets (WHU-Hi-LongKou, HongHu, HanChuan). Across all eight benchmarks, BCG-Former achieves the lowest inference latency while reaching state-of-the-art competitive accuracy, using only 0.10--0.23\,M parameters (up to 24$\times$ fewer than prior methods), thereby demonstrating a strong accuracy-efficiency trade-off across diverse benchmarks.

\end{itemize}

%

\section{Methodology}
Let the input hyperspectral patch be \(\mathbf{X} \in \mathbb{R}^{B \times C \times H \times W}\), where \(B\) is the batch size, \(C\) is the number of spectral bands, and \(H \times W \) is the patch size. The overall architecture consists of five main stages as follows:
(1) CNN stem for initial spatial feature extraction, (2) Band-Contextual Gating, (3) token fusion and positional encoding and single-pass band rotational positional encoding (RoPE), (4) linear spectral-spatial attention, and (5) the final classification head.

\subsection{CNN Stem}
 A two-layer CNN stem extracts local spatial features while preserving spectral information: \(\mathbf{F} = \text{Stem}(\mathbf{X})\),
where \(\text{Stem}(\cdot)\) comprises two \(3 \times 3\) convolutions (first with \(C \to  D \), where $D$ is the embedding dimension, followed by a second depthwise convolution), each followed by Batch Normalization and Sigmoid Linear Unit (SiLU) activation. The output \(\mathbf{F} \in \mathbb{R}^{B \times D \times H \times W}\) serves as the base spatial features.

\subsection{Band-Contextual Gating (BCG)}
To adaptively emphasize informative spectral features in the embedded space using local channel context, we introduce Band-Contextual Gating (BCG). First, global average pooling produces a spectral descriptor:
\begin{equation}
\mathbf{s} = \frac{1}{H \times W} \sum_{h=1}^{H} \sum_{w=1}^{W} \mathbf{F}_{:, :, h, w} \in \mathbb{R}^{B \times D}.
\end{equation}

Following this, a 1D convolution with kernel size 7 is applied along the spectral dimension (after transposing \(\mathbf{s}\) to shape \(\mathbb{R}^{B \times 1 \times D}\)) to capture local band context $\mathbf{s}'$, followed by a two-layer bottleneck to obtain the logits $\mathbf{L}$:
\begin{equation}
\mathbf{s}' = \text{Conv1D}(\mathbf{s}^\top) \in \mathbb{R}^{B \times D},
\quad
\mathbf{L} = \mathbf{W}_2 \cdot \text{SiLU}(\mathbf{W}_1 \mathbf{s}'),
\end{equation}
where \(\mathbf{W}_1 \in \mathbb{R}^{D \times (D/4)}\) and \(\mathbf{W}_2 \in \mathbb{R}^{(D/4) \times D}\).

An exponentially learnable temperature parameter \(\tau = \exp(\theta)\) (with \(\theta\) clamped so \(\tau \leq 10\)) sharpens the sigmoid gating $\sigma(\cdot)$. The gating vector is then reshaped to \(\mathbf{g} \in \mathbb{R}^{B \times D \times 1 \times 1}\) for broadcasting, to obtain the gated features $F'$:
\begin{equation}
\mathbf{g} = \sigma(\mathbf{L} \cdot \tau) \in \mathbb{R}^{B \times D},
\quad
\mathbf{F}' = \mathbf{F} \odot \mathbf{g}.
\end{equation}
This mechanism performs band-informed channel gating in the learned embedding space (dimension $  D  $), extending Squeeze-and-Excitation concepts to hyperspectral imagery by leveraging local spectral context along the embedded channels with temperature sharpening.

\subsection{Spectral Summary Token and Token Construction}
 The $\mathbf{F}'$ is then used to create the learnable spectral summary tokens that bridge spectral and spatial information. The summary is computed as:
\begin{equation}
\mathbf{s}_{\text{sum}} = \frac{1}{H \times W} \sum_{h,w} \mathbf{F}'_{:, :, h, w} \in \mathbb{R}^{B \times D}
\end{equation}
\begin{equation}
\mathbf{t}_{\text{spec}} = \mathbf{p}_{\text{spec}} + \mathbf{s}_{\text{sum}} \in \mathbb{R}^{B \times 1 \times D},
\end{equation}
where \(\mathbf{p}_{\text{spec}} \in \mathbb{R}^{1 \times 1 \times D}\) is a learnable parameter. This spectral summary token acts as a compact global spectral prior, allowing direct interaction between aggregated spectral information and local spatial tokens within the attention mechanism. By injecting this learned spectral representation, the model establishes a stronger spectral-spatial bridge compared to purely spatial tokenization approaches.

 Spatial tokens are obtained by flattening the gated features:
\begin{equation}
\mathbf{T}_{\text{spat}} = \text{Flatten}(\mathbf{F}')^\top \in \mathbb{R}^{B \times N \times D},
\end{equation}
with $N=H \times W=25$. The full token sequence is the concatenation of the spectral summary token and spatial tokens, plus a learnable positional embedding:
\begin{equation}
\mathbf{T}_0 = \text{Concat}(\mathbf{t}_{\text{spec}}, \mathbf{T}_{\text{spat}}) + \mathbf{P} \in \mathbb{R}^{B \times (1+N) \times D},
\end{equation}
where \(\mathbf{P} \in \mathbb{R}^{1 \times (1+N) \times D}\) is a learnable position parameter. Dropout is applied after addition.
After adding the learnable positional embedding, we apply a single-pass Band-RoPE to the entire token sequence \(\mathbf{T}_0\). Unlike conventional RoPE~\cite{rope} that is typically applied inside each attention layer on queries and keys, our implementation applies Band-RoPE once to all tokens before entering the Transformer blocks. Specifically, Band-RoPE treats the spectral tokens as an ordered band sequence and applies rotary positional encoding along the spectral dimension, while the spectral summary token is assigned a dedicated global position index to preserve its role as a scene-level aggregation token. The resulting position-aware tokens are then directly used to compute queries, keys, and values in the linear attention. This design combines absolute positional information from the learnable embedding \(\mathbf{P}\) with relative positional encoding from Band-RoPE, providing richer positional awareness while maintaining computational efficiency through single-pass processing.

\subsection{Linear Spectral-Spatial Attention and Classification Head}
The model stacks three Transformer blocks using efficient linear attention with an ELU-based kernel, achieving linear complexity in the sequence length $  O(N)  $ (specifically $  O(ND^2)  $ overall, where $  D  $ is the embedding dimension). For each block, layer normalization is applied, followed by a linear projection to obtain queries, keys, and values:
\begin{equation}
[\mathbf{Q}, \mathbf{K}, \mathbf{V}] = \text{Linear}(\text{LayerNorm}(\mathbf{T})) \in \mathbb{R}^{B \times (1+N) \times 3D}.
\end{equation}

After reshaping to multi-head form (\(M=4\) heads, head dimension \(d = D/M\)), we apply the Exponential Linear Unit (ELU) kernel on Queries and Keys:
\begin{equation}
\mathbf{Q}' = \text{ELU}(\mathbf{Q}) + 1, \quad
\mathbf{K}' = \text{ELU}(\mathbf{K}) + 1.
\end{equation}
The keys are then normalized over the sequence dimension (\(N\)):
\begin{equation}
\mathbf{K}'' = \frac{\mathbf{K}'}{\sum_{i=1}^{N} \mathbf{K}'_{:,i,:} + \epsilon},
\end{equation}
where the summation is performed along the token (sequence) dimension. The context matrix and output are computed as:
\begin{equation}
\mathbf{C} = (\mathbf{K}'')^\top \mathbf{V}, \quad
\mathbf{O} = \mathbf{Q}' \mathbf{C}.
\end{equation}

 The output is then passed through a projection and residual connection, each block also includes a standard MLP (GELU, ratio 2.0) with residual connection and LayerNorm.
After \(L\) blocks, the final tokens \(\mathbf{T}_L\) are normalized. The classification logit is obtained by fusing the spectral token and the mean of spatial tokens:
\begin{equation}
\mathbf{z} = \mathbf{T}_L[:,0] + \text{Mean}(\mathbf{T}_L[:,1:]) \in \mathbb{R}^{B \times D},
\end{equation}
\begin{equation}
\mathbf{y} = \mathbf{W}_{\text{head}} \mathbf{z} \in \mathbb{R}^{B \times K},
\end{equation}
where \(K\) is the number of classes and \(\mathbf{W}_{\text{head}}\) is the classification head.
 The model is trained with cross-entropy loss. All weights are initialized with truncated normal distribution (\(\sigma=0.02\)).
This design achieves strong spectral-spatial fusion with low computational cost due to linear attention and single-pass RoPE, while the BCG and spectral summary token provide explicit spectral guidance.

\section{Experiments and Results}

\subsection{Experimental Setup}
All experiments were conducted on a workstation with an NVIDIA RTX A6000 GPU (48\,GB VRAM), AMD EPYC 9354 32-core CPU, and 755\,GB RAM. The proposed BCG-Former and all baselines were evaluated under identical hardware and software settings on eight hyperspectral benchmarks spanning airborne (Houston 2013 \cite{houston}, Houston 2018 \cite{houston18}, Pavia University \cite{paviauniversity}, Salinas \cite{salinas}, and Indian Pines \cite{indian_pines_1992}), and UAV-borne sensing scenarios (three UAV-borne datasets: WHU-Hi-LongKou \cite{whu_hi_datasets}, HongHu  \cite{whu_hi_datasets}, and HanChuan \cite{whu_hi_datasets}). 

For each dataset, $5 \times 5$
spatial patches were extracted using reflected padding at the image boundaries. Following recent HSI literature \cite{hit, ssftt}, we adopt a fixed 200 samples-per-class training protocol with spatial exclusion to prevent overlap between training and test patch centers. For Houston 2013, we use the official GRSS 2013 Data Fusion Contest fixed split (2,832 train / 12,197 test) \cite{houston}; for Houston 2018, no pixel-level official split is publicly available as the contest test ground truth was withheld during the competition, so we apply the standard 200 samples-per-class protocol consistent with recent literature \cite{houston18}; for all remaining six datasets, the same 200 samples-per-class protocol is applied. This class-balanced setup is preferred over percentage-based splits, particularly for imbalanced datasets such as Indian Pines where several minority classes contain fewer than 50 total labeled samples. All results are reported as mean $\pm$
standard deviation over five independent runs; for Houston 2013, run-to-run variance reflects model initialization stochasticity rather than data partitioning, consistent with the standard evaluation protocol for this dataset \cite{houston}, whereas for all remaining seven datasets, variance reflects both split and initialization randomness.

BCG-Former is configured with a 64-dimensional embedding size, 3 transformer blocks, 4 attention heads, and an MLP ratio of 2.0, resulting in 0.10-0.23\,M trainable parameters. It is trained for 50 epochs using AdamW with a learning rate of $3 \times 10^{-4}$, a weight decay of $10^{-2}$, cosine scheduling with 10\% warmup, a batch size of 32, gradient clipping (norm 1.0), and label smoothing of 0.1. All results are reported as mean $\pm$ standard deviation over five runs with different random seeds, and performance is evaluated using Overall Accuracy (OA), Average Accuracy (AA), Cohen{'}s Kappa ($\kappa$), inference latency, parameter count, and GFLOPs.

\begin{table*}[t]
\centering
\small
\caption{Comprehensive Performance and Efficiency Comparison on \textbf{Classical} (PU: Pavia University~\cite{paviauniversity}, SA: Salinas~\cite{salinas}, IN: Indian Pines~\cite{indian_pines_1992}, H13: Houston 13~\cite{houston}, and  H18: Houston 18~\cite{houston18}) and \textbf{UAV-borne} Datasets (HH: WHU-Hi-HongHu ~\cite{whu_hi_datasets}, HC: WHU-Hi-HanChuan~\cite{whu_hi_datasets}, and LK: WHU-Hi-LongKou~\cite{whu_hi_datasets}).}
\label{tab:main_comparison}

\begin{minipage}{\textwidth}
\centering
\small (a) Classification Performance (OA, AA, and Kappa)
\renewcommand{\arraystretch}{1.05}
\begin{tabular*}{\textwidth}{@{\extracolsep{\fill}}|l|ccccc|ccc|}
\hline
\textbf{Model} & \textbf{PU} & \textbf{SA} & \textbf{IN} & \textbf{H13} & \textbf{H18} & \textbf{HH} & \textbf{HC} & \textbf{LK} \\ \hline
\multicolumn{9}{|c|}{\textit{Overall Accuracy (OA \%)}} \\ \hline
HiT~\cite{hit}           & \underline{98.00} & \underline{94.30} & 95.40 & 98.62 & 89.30 & \underline{92.99} & \textbf{92.06} & 98.07 \\
HybridSN~\cite{hybridsn} & 97.70 & 93.24 & 86.99 & 98.03 & 87.29 & 91.17 & 90.23 & 97.49 \\
SpectralMamba~\cite{spectralmamba} & 97.47 & 94.12 & \underline{95.76} & 98.77 & 90.28 & 91.24 & 90.40 & \underline{98.32} \\
SSFTT~\cite{ssftt}       & 97.20 & 93.71 & 84.88 & 98.04 & 87.24 & 90.44 & 88.90 & 97.22 \\
Swin-HSI~\cite{swinhsi}  & 97.71 & 94.10 & 95.35 & \underline{98.86} & \underline{91.10} & 90.99 & 91.51 & 98.15 \\ 
\textbf{BCG-Former (Ours)} & \textbf{99.08} & \textbf{96.44} & \textbf{96.07} & \textbf{99.49} & \textbf{91.51} & \textbf{93.86} & \underline{91.87} & \textbf{99.43} \\ \hline
\multicolumn{9}{|c|}{\textit{Average Accuracy (AA \%)}} \\ \hline
HiT~\cite{hit}           & \underline{98.13} & \underline{97.26} & 81.44 & 98.64 & \underline{89.21} & \textbf{93.93} & \textbf{91.46} & 97.77 \\
HybridSN~\cite{hybridsn} & 97.62 & 96.61 & 81.09 & 98.09 & 82.50 & 91.48 & 89.84 & 97.28 \\
SpectralMamba~\cite{spectralmamba} & 97.27 & 97.23 & 80.40 & 98.79 & 85.65 & 91.31 & 89.64 & \underline{98.25} \\
SSFTT~\cite{ssftt}       & 97.03 & 96.84 & 82.86 & 98.05 & 86.06 & 90.08 & 86.71 & 96.98 \\
Swin-HSI~\cite{swinhsi}  & 97.51 & 97.17 & \textbf{93.00} & \underline{98.86} & \textbf{89.65} & 91.36 & \underline{90.91} & 97.87 \\
\textbf{BCG-Former (Ours)} & \textbf{99.02} & \textbf{99.87} & \underline{85.16} & \textbf{99.52} & 87.40 & \underline{93.39} & 90.76 & \textbf{99.16} \\ \hline
\multicolumn{9}{|c|}{\textit{Kappa Coefficient}} \\ \hline
HiT~\cite{hit}           & 0.977 & 0.938 & 0.949 & 0.986 & 0.831 & 0.914 & \textbf{0.909} & 0.978 \\
HybridSN~\cite{hybridsn} & 0.973 & 0.926 & 0.853 & 0.979 & 0.793 & 0.892 & 0.888 & 0.970 \\
SpectralMamba~\cite{spectralmamba} & 0.969 & 0.936 & 0.951 & 0.987 & 0.843 & 0.892 & 0.890 & 0.981 \\
SSFTT~\cite{ssftt}       & 0.966 & 0.931 & 0.829 & 0.979 & 0.794 & 0.882 & 0.872 & 0.967 \\
Swin-HSI~\cite{swinhsi}  & 0.973 & 0.935 & 0.948 & 0.988 & \textbf{0.858} & 0.890 & 0.900 & 0.979 \\
\textbf{BCG-Former (Ours)} & \textbf{0.994} & \textbf{0.958} & \textbf{0.952} & \textbf{0.994} & \underline{0.844} & \textbf{0.918} & \underline{0.902} & \textbf{0.991} \\ \hline
\end{tabular*}
\end{minipage}

\vspace{1.0em}

\begin{minipage}{\textwidth}
\centering
\small (b) Computational Efficiency (Complexity, Speed, and Size)
\renewcommand{\arraystretch}{1.05}
\begin{tabular*}{\textwidth}{@{\extracolsep{\fill}}|l|ccccc|ccc|}
\hline
\textbf{Model} & \textbf{PU} & \textbf{SA} & \textbf{IN} & \textbf{H13} & \textbf{H18} & \textbf{HH} & \textbf{HC} & \textbf{LK} \\ \hline
\multicolumn{9}{|c|}{\textit{Computational Complexity (GFLOPs)}} \\ \hline
HiT~\cite{hit}           & 0.0213 & 0.0432 & 0.0425 & 0.0114 & 0.0114 & 0.0515 & 0.0523 & 0.0515 \\
HybridSN~\cite{hybridsn} & 0.2004 & 0.4348 & 0.4271 & 0.0939 & 0.0939 & 0.5239 & 0.5317 & 0.5239 \\
SpectralMamba~\cite{spectralmamba} & \textbf{0.0039} & \textbf{0.0046} & \textbf{0.0046} & \textbf{0.0035} & \textbf{0.0035} & \textbf{0.0049} & \textbf{0.0049} & \textbf{0.0049} \\
SSFTT~\cite{ssftt}       & 0.1157 & 0.1468 & 0.1458 & 0.1015 & 0.1015 & 0.1586 & 0.1597 & 0.1586 \\
Swin-HSI~\cite{swinhsi}  & 0.0089 & \underline{0.0093} & \underline{0.0093} & 0.0088 & 0.0088 & \underline{0.0095} & \underline{0.0095} & \underline{0.0095} \\
\textbf{BCG-Former (Ours)} & \underline{0.0065} & 0.0100 & 0.0099 & \underline{0.0049} & \underline{0.0049} & 0.0113 & 0.0114 & 0.0113 \\ \hline
\multicolumn{9}{|c|}{\textit{Inference Speed (Latency in ms )}} \\ \hline
HiT~\cite{hit}           & \underline{1.1916} & 1.7716 & 1.7538 & 1.1848 & 1.1805 & 2.0286 & 2.0522 & 2.0356 \\
HybridSN~\cite{hybridsn} & 2.0320 & 4.2525 & 4.1883 & \underline{0.9807} & \underline{0.9824} & 5.0771 & 5.1470 & 5.0691 \\
SpectralMamba~\cite{spectralmamba} & 1.4445 & 1.3880 & 1.3868 & 1.4413 & 1.4553 & 1.4042 & 1.3874 & 1.4118 \\
SSFTT~\cite{ssftt}       & 5.5391 & 5.8103 & 5.8601 & 5.5734 & 5.4998 & 5.9280 & 5.8626 & 5.9404 \\
Swin-HSI~\cite{swinhsi}  & 1.3976 & \underline{1.3855} & \underline{1.3512} & 1.3358 & 1.4073 & \underline{1.3548} & \underline{1.3853} & \underline{1.3764} \\
\textbf{BCG-Former (Ours)} & \textbf{0.9398} & \textbf{0.9352} & \textbf{0.9347} & \textbf{0.9271} & \textbf{0.9528} & \textbf{0.9115} & \textbf{0.9371} & \textbf{0.9366} \\ \hline
\multicolumn{9}{|c|}{\textit{Network Size (Parameters in Mil)}} \\ \hline
HiT~\cite{hit}           & 0.33 & 0.57 & 0.57 & 0.21 & 0.21 & 0.67 & 0.68 & 0.67 \\
HybridSN~\cite{hybridsn} & 2.36 & 4.59 & 4.52 & 1.35 & 1.35 & 5.44 & 5.52 & 5.44 \\
SpectralMamba~\cite{spectralmamba} & 0.15 & \textbf{0.16} & \textbf{0.16} & 0.14 & 0.14 &\textbf{ 0.17} &\textbf{ 0.17} &\textbf{ 0.17} \\
SSFTT~\cite{ssftt}       & 1.11 & 1.66 & 1.65 & 0.85 & 0.85 & 1.88 & 1.90 & 1.87 \\
Swin-HSI~\cite{swinhsi}  & 0.58 & 0.58 & 0.58 & 0.57 & 0.57 & 0.59 & 0.59 & 0.59 \\
\textbf{BCG-Former (Ours)} & \textbf{0.13} & \underline{0.20} & \underline{0.20} & \textbf{0.10} & \textbf{0.10} & \underline{0.23} & \underline{0.23} & \underline{0.23} \\ \hline
\end{tabular*}
\end{minipage}

\end{table*}

\begin{figure*}[t]
    \centering
    \begin{subfigure}[b]{0.242\textwidth}
        \fbox{\includegraphics[width=\textwidth]{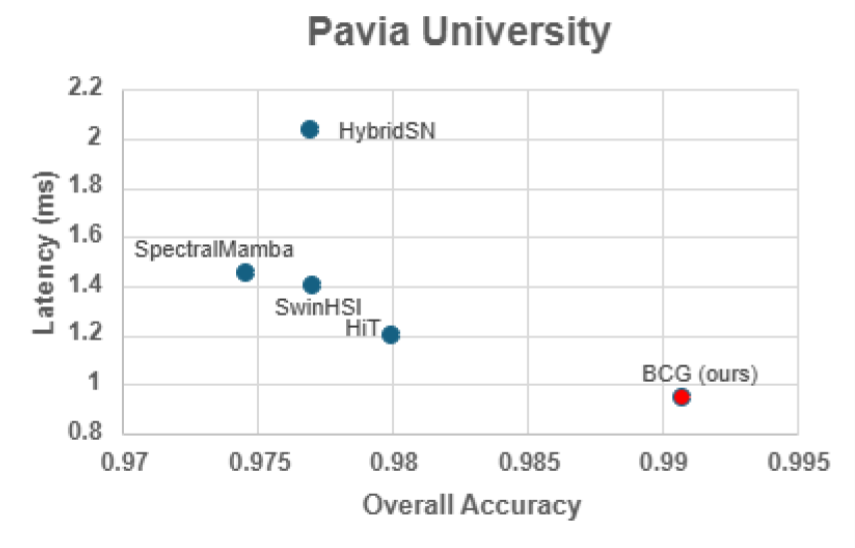}}
    \end{subfigure}
    \hfill
    \begin{subfigure}[b]{0.239\textwidth}
        \fbox{\includegraphics[width=\textwidth]{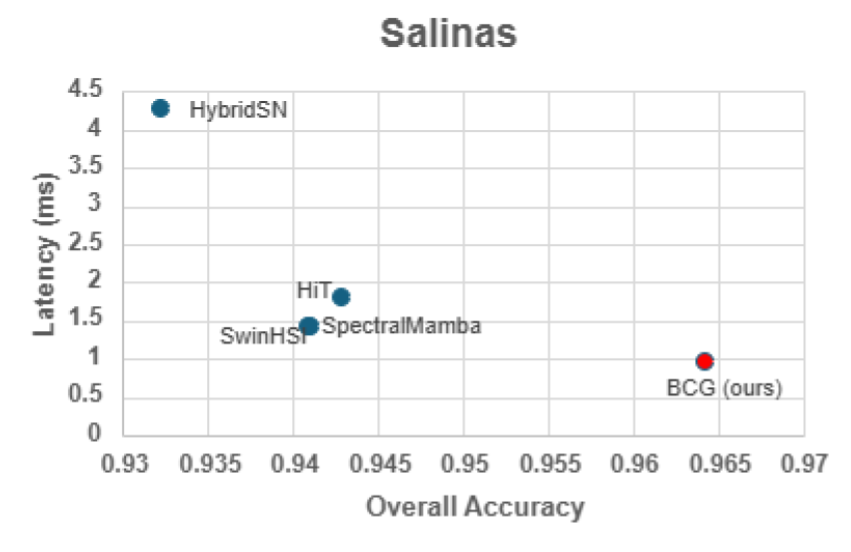}}
    \end{subfigure}
    \hfill
    \begin{subfigure}[b]{0.2395\textwidth}
        \fbox{\includegraphics[width=\textwidth]{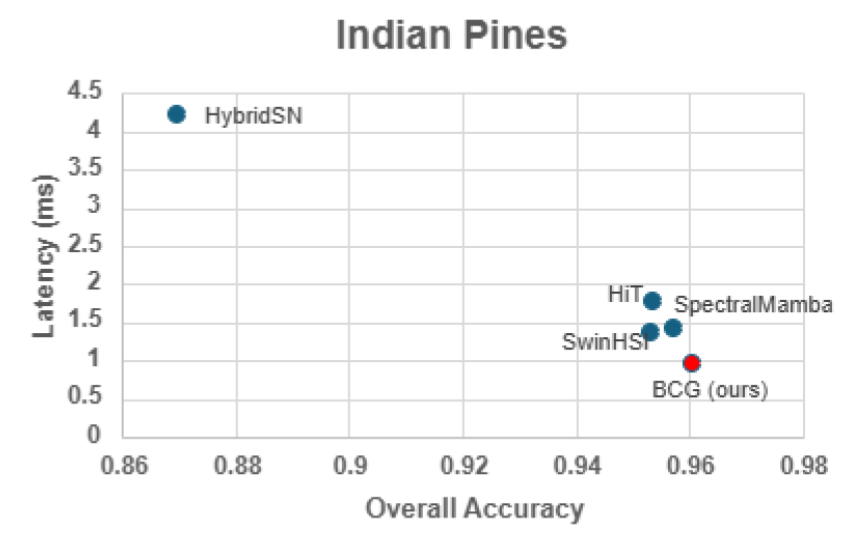}}
    \end{subfigure}
    \hfill
    \begin{subfigure}[b]{0.2375\textwidth}
        \fbox{\includegraphics[width=\textwidth]{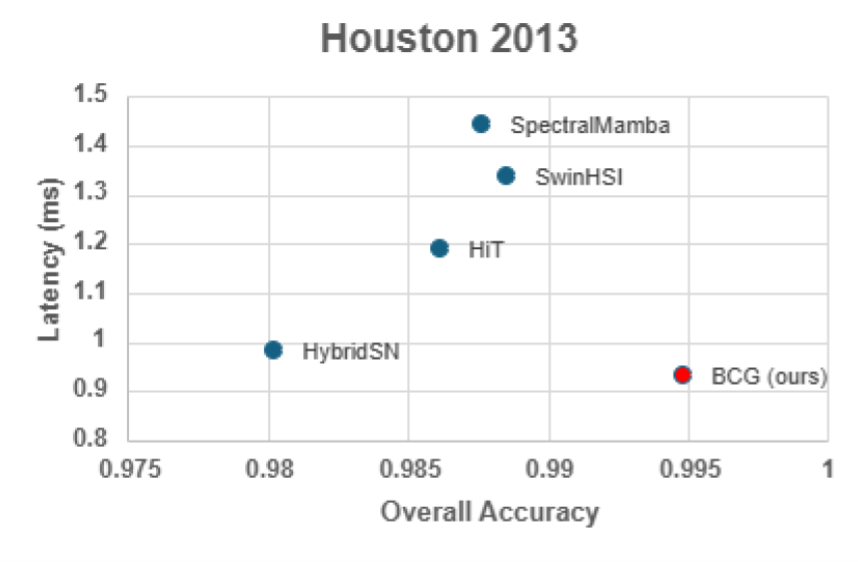}}
    \end{subfigure}
    \\[0.5em]
    \begin{subfigure}[b]{0.24\textwidth}
        \fbox{\includegraphics[width=\textwidth]{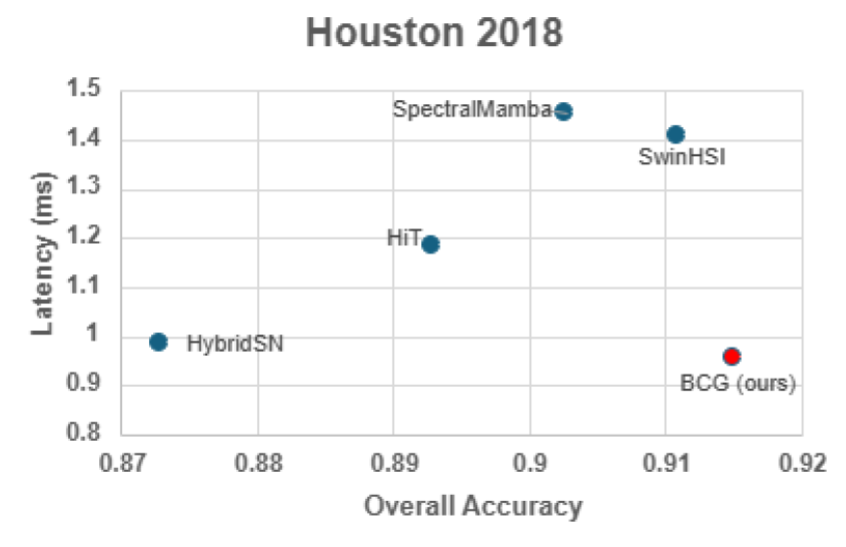}}
    \end{subfigure}
    \hfill
    \begin{subfigure}[b]{0.239\textwidth}
        \fbox{\includegraphics[width=\textwidth]{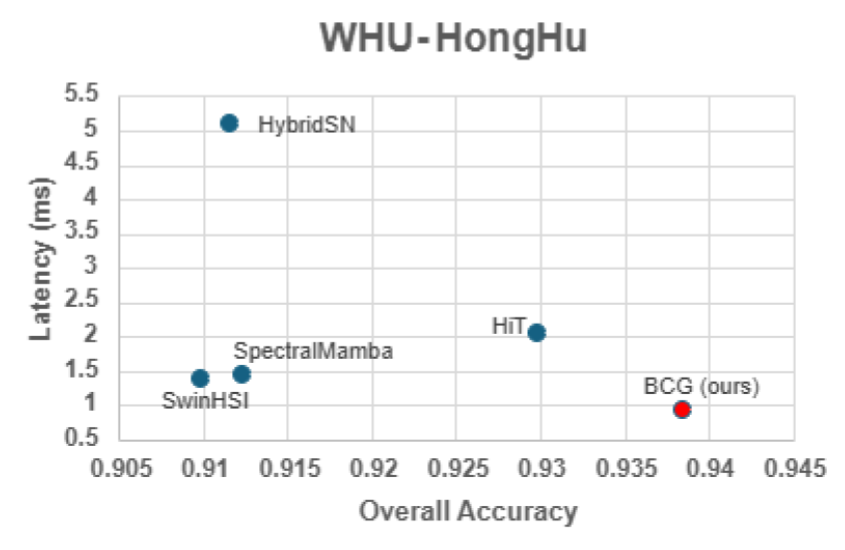}}
    \end{subfigure}
    \hfill
    \begin{subfigure}[b]{0.239\textwidth}
        \fbox{\includegraphics[width=\textwidth]{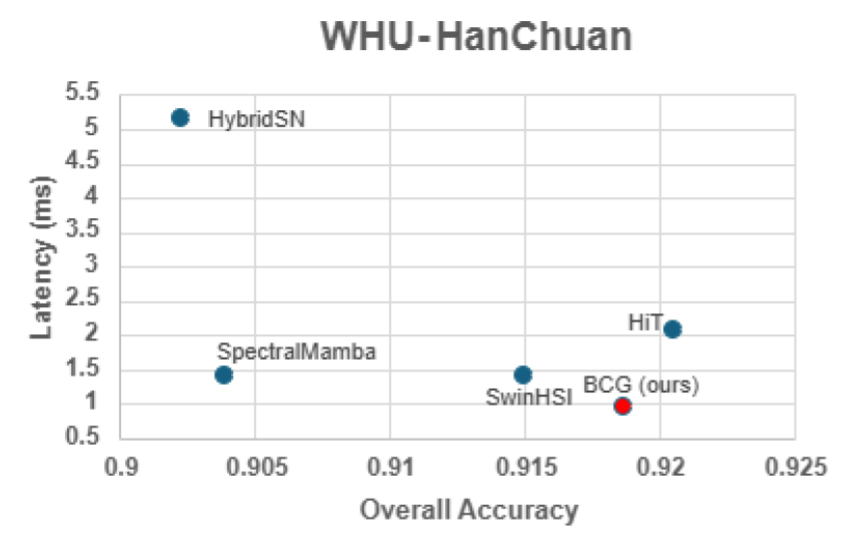}}
    \end{subfigure}
    \hfill
    \begin{subfigure}[b]{0.24\textwidth}
        \fbox{\includegraphics[width=\textwidth]{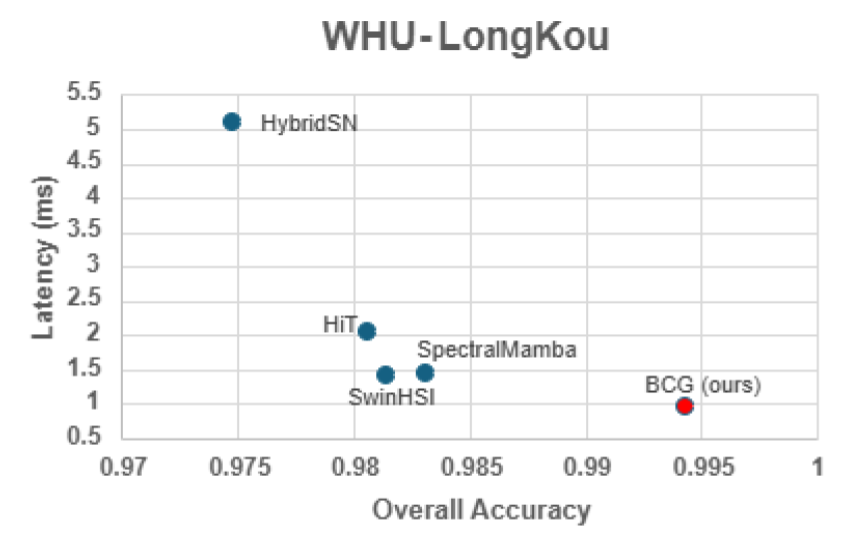}}
    \end{subfigure}
\caption{Accuracy-latency Pareto frontier plots across all eight benchmark datasets. BCG-Former (Ours) consistently resides on or near the Pareto frontier, achieving competitive or state-of-the-art accuracy at the lowest inference latency. SSFTT is excluded from the plots as its substantially higher inference latency (5.5-5.9\,ms) renders it a clear outlier and compresses the visualization of the remaining methods.}
    \label{fig:pareto_plots}
\end{figure*}

\subsection{Classification Performance}
\begin{figure*}[t]
\centering
\includegraphics[scale=0.4]{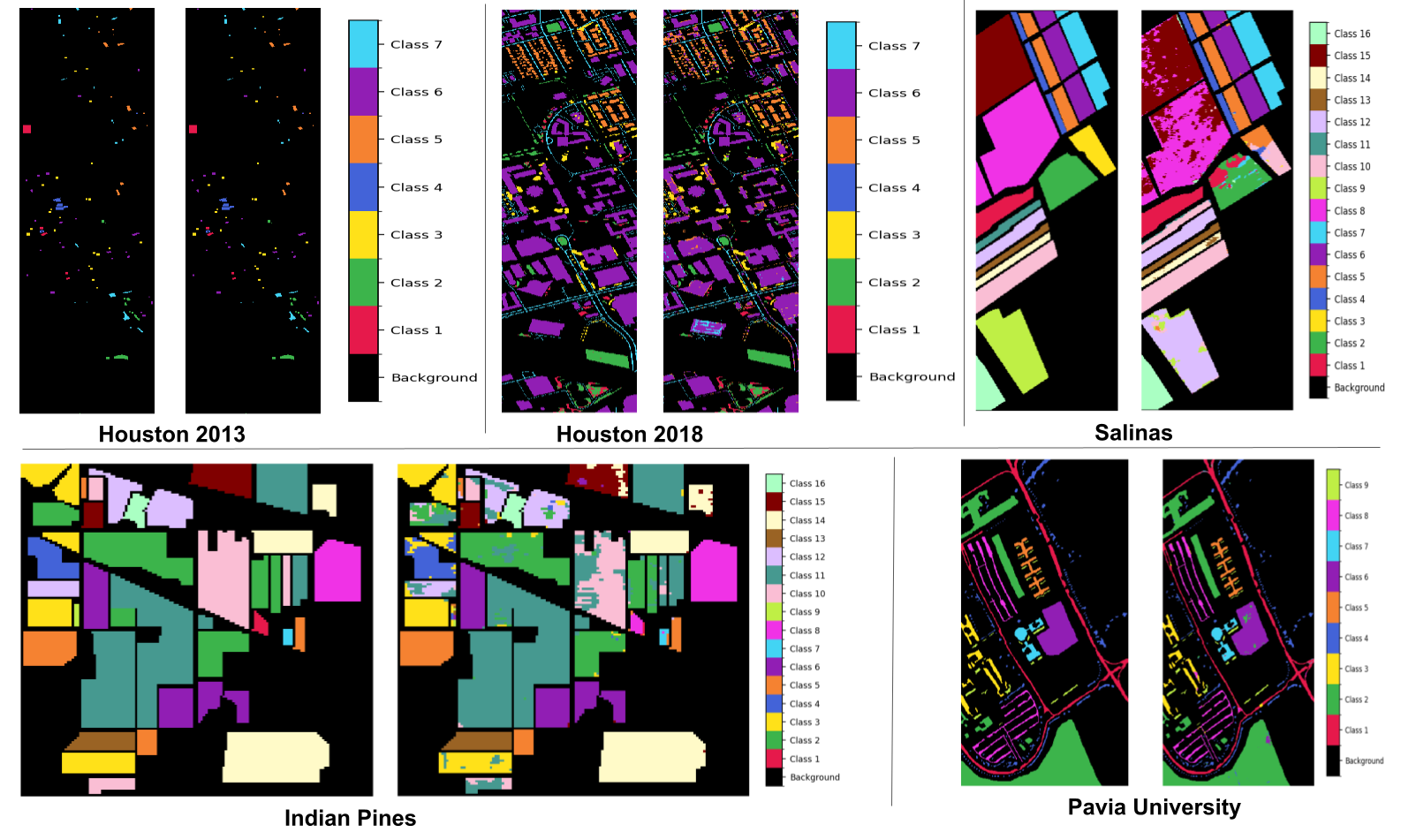}
\caption{Predicted classification maps generated by BCG-Former on airborne datasets (Pavia University, Salinas, Indian Pines, Houston 2013, and Houston 2018). For each dataset, the ground truth (GT) map is shown on the left and the corresponding predicted map on the right.}
\label{fig:airborne_maps}
\end{figure*}

\begin{figure*}[t]
\centering
\includegraphics[scale=0.35]{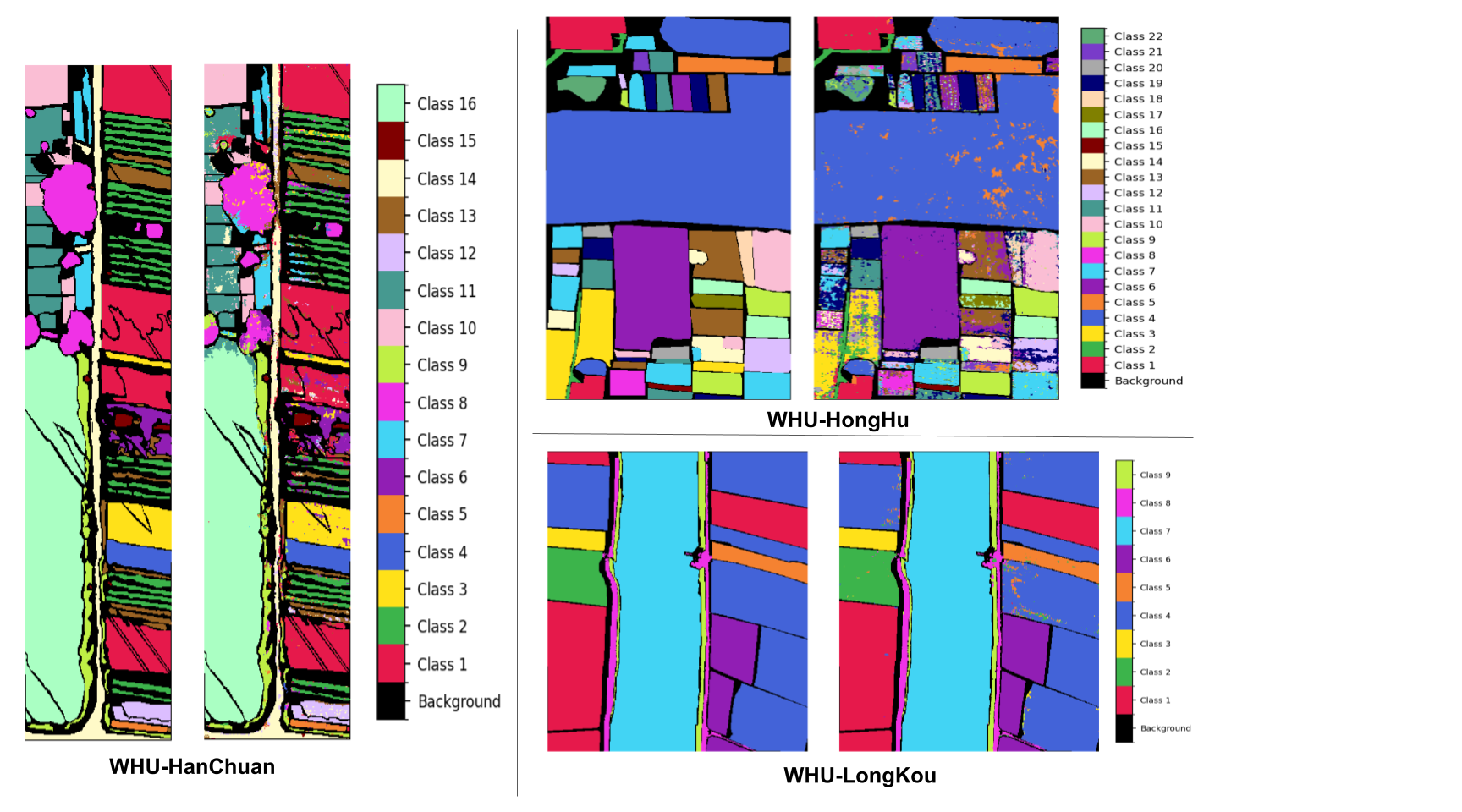}
\caption{Predicted classification maps generated by BCG-Former on UAV-borne datasets (WHU-Hi-HongHu, WHU-Hi-HanChuan, and WHU-Hi-LongKou). For each dataset, the ground truth (GT) map is shown on the left and the corresponding predicted map on the right.}
\label{fig:uav_maps}
\end{figure*}

As shown in Table~\ref{tab:main_comparison}(a), BCG-Former achieves the highest overall accuracy on seven of eight benchmarks. On airborne datasets, it records 99.08\%, 96.44\%, 99.49\%, and 91.51\% OA on PU, SA, H13, and H18, consistently outperforming all CNN-, Transformer-, and Mamba-based competitors. On Indian Pines, BCG-Former achieves the best OA of 96.07\%, confirming strong discrimination despite high spectral similarity among crop classes, and significant intra-class variability. Although AA on Indian Pines and Houston 2018 (85.16\% and 87.40\%) remains below the strongest baseline (93.00\% and 89.65\%) due to minority-class imbalance, OA and Kappa remain highly competitive, reflecting strong overall discrimination. Across UAV-borne benchmarks, BCG-Former achieves 93.86\%, 91.87\%, and 99.43\% OA on HH, HC, and LK respectively, the highest on HH and LK. Although AA on HongHU and HanChuan remains marginally below HiT (by 0.54\% and 0.70\% respectively), OA and Kappa remain leading on HH and LK, reflecting strong overall discrimination; the narrow AA gap on these UAV-borne benchmarks is consistent with the dense spatial overlap and high spectral complexity characteristic of low-altitude urban sensing. All results are reported as mean $\pm$ standard deviation over five independent runs. BCG-Former demonstrates stable optimization across all benchmarks, with OA standard deviations of $\pm$0.26\%, $\pm$0.13\%, $\pm$0.27\%, $\pm$0.23\%, $\pm$0.98\%, $\pm$0.24\%, $\pm$0.57\%, and $\pm$0.11\% on PU, SA, IN, H13, H18, HH, HC, and LK respectively, indicating that the proposed method provides a strong and reliable inductive bias across diverse scene types. The highest variance is observed on Houston 2018 ($\pm0.98\%$), attributable to the absence of a fixed official split, where both data partitioning and model initialisation contribute to run-to-run variation. Figs.~\ref{fig:airborne_maps} and \ref{fig:uav_maps} represent the classification maps of both airborne and UAV-borne datasets, respectively. 

\subsection{Computational Efficiency}
Table~\ref{tab:main_comparison}(b) presents a detailed comparison of computational efficiency across all eight benchmarks. BCG-Former achieves low computational complexity, with GFLOPs ranging from 0.0049 to 0.0114, while consistently delivering the lowest inference latency (0.91--0.95\,ms) on all datasets compared to other methods. Notably, GFLOPs and latency are not always directly correlated, as GFLOPs measure theoretical computation, whereas latency reflects actual hardware execution time. Although SpectralMamba reports the lowest GFLOPs, its sequential scanning operation limits GPU parallelism and results in higher latency; this dissociation underscores why wall-clock inference time is a more reliable efficiency metric than arithmetic operation count for deployment-oriented evaluation. HybridSN attains reasonably competitive latency by leveraging optimized 3D convolutions; however, this comes at the expense of significantly larger model sizes (up to 5.52\,M parameters, almost $24\times$ that of BCG-Former), and its parameter count grows further as spectral dimensionality increases. 

BCG-Former avoids these trade-offs by utilizing fully parallelizable operations: depthwise convolution, single-pass Band-RoPE, and ELU-kernel linear attention. This formulation further reduces self-attention complexity from $\mathcal{O}(N^{2}D)$ to $\mathcal{O}(ND^{2})$. While the token count remains small at $N=26$ for patch-based classification, this architecture is extensible to larger spatial contexts without modification. Consequently, BCG-Former maintains a compact parameter footprint of 0.10--0.23\,M (second only to SpectralMamba on five datasets) while achieving the lowest latency across all eight benchmarks without requiring dataset-specific architectural modifications. Considered jointly across latency, parameters, and classification accuracy, BCG-Former is the only model that simultaneously achieves sub-millisecond inference, a sub-0.24\,M parameter count, and state-of-the-art accuracy on seven of eight benchmarks. \textit{This balanced profile firmly places BCG-Former on the Pareto frontier of accuracy, model size, and inference efficiency among existing CNN-, Transformer-, and Mamba-based HSI classification methods, as illustrated in Fig.~\ref{fig:pareto_plots}.}

\begin{table*}[t]
\centering
\small
\caption{Comprehensive Ablation Study on \textbf{Classical} (PU: Pavia University~\cite{paviauniversity}, SA: Salinas~\cite{salinas}, IN: Indian Pines~\cite{indian_pines_1992}, H13: Houston 13~\cite{houston}, and H18: Houston 18~\cite{houston18}) and \textbf{UAV-borne} Datasets (HH: WHU-Hi-HongHu ~\cite{whu_hi_datasets}, HC: WHU-Hi-HanChuan~\cite{whu_hi_datasets}, and LK: WHU-Hi-LongKou~\cite{whu_hi_datasets}).}
\label{tab:main_ablation}

\begin{subtable}{\textwidth}
\centering
\caption{Classification Performance Ablation (OA, AA, and Kappa)}
\renewcommand{\arraystretch}{1.05}
\begin{tabular*}{\textwidth}{@{\extracolsep{\fill}}|l|ccccc|ccc|}
\hline
\textbf{Model Variant} & \textbf{PU} & \textbf{SA} & \textbf{IN} & \textbf{H13} & \textbf{H18} & \textbf{HH} & \textbf{HC} & \textbf{LK} \\ \hline
\multicolumn{9}{|c|}{\textit{Overall Accuracy (OA \%)}} \\ \hline
w/o BCG                    & 97.94 & 94.59 & 94.58 & 99.33 & 90.18 & 91.72 & 90.13 & 98.84 \\
w/o Linear Spec-Spat Attn  & 97.93 & 94.48 & 94.15 & 99.33 & 90.15 & 91.22 & 89.51 & 98.82 \\
w/o Fusion \& Pos Enc      & 97.82 & 94.64 & 93.41 & 99.47 & 89.96 & 91.73 & 90.29 & 98.88 \\
\textbf{BCG-Former (Full)} & \textbf{99.08} & \textbf{96.44} & \textbf{96.07} & \textbf{99.49} & \textbf{91.51} & \textbf{93.86} & \textbf{91.87} & \textbf{99.43} \\ \hline
\multicolumn{9}{|c|}{\textit{Average Accuracy (AA \%)}} \\ \hline
w/o BCG                    & 98.00 & 97.93 & 84.08 & 99.37 & 86.97 & 91.70 & 89.14 & 98.32 \\
w/o Linear Spec-Spat Attn  & 97.96 & 97.89 & 79.84 & 99.36 & 82.18 & 90.66 & 88.20 & 98.65 \\
w/o Fusion \& Pos Enc      & 97.95 & 97.97 & 76.15 & 99.50 & 84.75 & 91.36 & 89.18 & 98.48 \\
\textbf{BCG-Former (Full)} & \textbf{99.02} & \textbf{99.87} & \textbf{85.16} & \textbf{99.52} & \textbf{87.40} & \textbf{93.39} & \textbf{90.76} & \textbf{99.16} \\ \hline
\multicolumn{9}{|c|}{\textit{Kappa Coefficient}} \\ \hline
w/o BCG                    & 0.972 & 0.940 & 0.938 & 0.992 & 0.836 & 0.896 & 0.885 & 0.985 \\
w/o Linear Spec-Spat Attn  & 0.972 & 0.938 & 0.933 & 0.992 & 0.837 & 0.890 & 0.878 & 0.985 \\
w/o Fusion \& Pos Enc      & 0.971 & 0.940 & 0.925 & 0.994 & 0.834 & 0.896 & 0.887 & 0.985 \\
\textbf{BCG-Former (Full)} & \textbf{0.994} & \textbf{0.958} & \textbf{0.952} & \textbf{0.994} & \textbf{0.844} & \textbf{0.918} & \textbf{0.902} & \textbf{0.991} \\ \hline
\end{tabular*}
\label{tab:ablation_perf}
\end{subtable}

\vspace{0.5em}

\begin{subtable}{\textwidth}
\centering
\caption{Computational Efficiency Ablation (Complexity, Speed, and Size)}
\renewcommand{\arraystretch}{1.05}
\begin{tabular*}{\textwidth}{@{\extracolsep{\fill}}|l|ccccc|ccc|}
\hline
\textbf{Model Variant} & \textbf{PU} & \textbf{SA} & \textbf{IN} & \textbf{H13} & \textbf{H18} & \textbf{HH} & \textbf{HC} & \textbf{LK} \\ \hline
\multicolumn{9}{|c|}{\textit{Computational Complexity (GFLOPs $\downarrow$)}} \\ \hline
w/o BCG                    & 0.0065 & 0.0100 & 0.0099 & 0.0049 & 0.0049 & 0.0113 & 0.0114 & 0.0113 \\
w/o Linear Spec-Spat Attn  & 0.0065 & 0.0100 & 0.0099 & 0.0049 & 0.0049 & 0.0113 & 0.0114 & 0.0113 \\
w/o Fusion \& Pos Enc & \textbf{0.0064} & \textbf{0.0099} & \textbf{0.0097} & \textbf{0.0048} & \textbf{0.0048} & \textbf{0.0112} & \textbf{0.0113} & \textbf{0.0112} \\
BCG-Former (Full) & \underline{0.0065} & \underline{0.0100} & \underline{0.0099} & \underline{0.0049} & \underline{0.0049} & \underline{0.0113} & \underline{0.0114} & \underline{0.0113} \\ \hline
\multicolumn{9}{|c|}{\textit{Inference Speed (Latency in ms $\downarrow$)}} \\ \hline
w/o BCG                    & \textbf{0.9317} & 0.9799 & \underline{0.9408} & \underline{0.9811} & \underline{0.9504} & 0.9677 & \textbf{0.8846} & \textbf{0.8594} \\
w/o Linear Spec-Spat Attn  & 1.0441 & \underline{0.9791} & 0.9683 & 0.9947 & 1.0054 & 1.0127 & \underline{0.9145} & \underline{0.8886} \\
w/o Fusion \& Pos Enc      & 1.0889 & 1.0410 & 1.0335 & 1.0265 & 1.0924 & \underline{0.9340} & 0.9746 & 0.9489 \\
\textbf{BCG-Former (Full)} & \underline{0.9398} & \textbf{0.9352} & \textbf{0.9347} & \textbf{0.9271} & \textbf{0.9528} & \textbf{0.9115} & 0.9371 & 0.9366 \\ \hline
\multicolumn{9}{|c|}{\textit{Network Size (Parameters in M $\downarrow$)}} \\ \hline
w/o BCG                    & \textbf{0.130} & \textbf{0.199} & \textbf{0.199} & \textbf{0.099} & \textbf{0.099} & \textbf{0.218} & \textbf{0.218} & \textbf{0.218} \\
w/o Linear Spec-Spat Attn  & 0.130 & 0.200 & 0.200 & 0.100 & 0.100 & 0.230 & 0.230 & 0.230 \\
w/o Fusion \& Pos Enc      & 0.130 & 0.200 & 0.200 & 0.100 & 0.100 & 0.230 & 0.230 & 0.230 \\
\textbf{BCG-Former (Full)} & \underline{0.130} & \underline{0.200} & \underline{0.200} & \underline{0.100} & \underline{0.100} & \underline{0.230} & \underline{0.230} & \underline{0.230}\\ \hline
\end{tabular*}
\label{tab:ablation_eff}
\end{subtable}

\end{table*}

\subsection{Ablation Study}
To evaluate the contribution of each component, we performed ablation studies by systematically removing: (1)~Band-Contextual Gating (BCG), (2)~the spectral summary token with spectral-spatial fusion, and (3)~linear attention, which was replaced by standard softmax attention. Parameter count remains nearly unchanged across all variants ($\sim$0.13\,M for Pavia University), since the novel components contribute negligible parameters by design, BCG adds only $\sim$2,100 parameters through a lightweight 1D convolution and two-layer MLP, the spectral summary token adds only 64 learnable parameters, and replacing linear with softmax attention changes zero parameters as both share identical QKV projections. This confirms that BCG-Former's accuracy gains arise from architectural inductive bias rather than parameter scaling. GFLOPs similarly remain near-identical across variants, since the dominant arithmetic cost lies in the shared transformer blocks; the marginal reduction in the \textit{w/o Fusion \& Pos Enc} variant simply reflects the removal of one token from the sequence length.

A notable observation is that BCG-Former (Full) achieves lower inference latency than several ablated variants on six of eight datasets, despite incorporating all components. This is explained by three complementary effects: BCG suppresses redundant spectral channels before tokenization, reducing uninformative activation flow through the transformer; the spectral summary token concentrates cross-modal reasoning into a single dedicated token, reducing the representational burden on spatial tokens; and linear attention avoids materializing the full $N \times N$ matrix and eliminates softmax normalisation overhead, enabling GPU kernel fusion that softmax prevents. Removing components therefore does not reduce latency, the full model is faster because its components produce a more hardware-efficient computation graph.

Table~\ref{tab:main_ablation} summarizes accuracy impact across all eight benchmarks. The full model consistently achieves the best accuracy-efficiency trade-off, confirming complementary and mutually beneficial interactions among all components. Removing BCG causes the largest degradation, OA drops by up to 1.49\% on airborne and 2.14\% on UAV-borne datasets, delivered through only $\sim$2,100 parameters, making it the highest accuracy-per-parameter component in the architecture. Removing linear attention produces the second-largest drop, particularly on Indian Pines and Houston 2018, while disabling the spectral summary token causes consistent decline on spectrally complex scenes. Across all ablation settings, BCG-Former (Full) achieves the highest accuracy while incurring lower latency than ablated counterparts on six of eight datasets, confirming that its Pareto-optimal position emerges from the synergistic interaction of three physically motivated, parameter-efficient components rather than any single design choice.

\section{Conclusion}
This paper presents \textbf{BCG-Former}, a lightweight CNN-Transformer designed to explicitly optimize the accuracy-latency Pareto frontier in hyperspectral image classification. Through three physically motivated components---Band-Contextual Gating, a spectral summary token, and single-pass Band-RoPE with linear spectral-spatial attention---BCG-Former achieves the lowest inference latency across all eight benchmark datasets while delivering state-of-the-art or highly competitive accuracy, using only 0.10--0.23\,M parameters without dataset-specific tuning. Ablation studies confirm that Band-Contextual Gating provides the largest individual contribution to spectral discrimination, while all components act synergistically to maintain Pareto optimality across diverse airborne and UAV-borne scenes. Overall, BCG-Former demonstrates that accuracy and efficiency are not mutually exclusive; carefully designed inductive biases aligned with the physical structure of hyperspectral data are sufficient to outperform larger and more complex state-of-the-art methods on both axes, establishing a strong efficiency-aware baseline for real-time geospatial remote sensing. Future work will explore hard band selection within the BCG framework, extension to multitemporal and multimodal HSI fusion, and adaptation to full-image inference where the linear attention formulation offers direct scalability advantages.

\bibliographystyle{IEEEtran}
\bibliography{ref}

@book{richards2013remote,
  author    = {J. A. Richards},
  title     = {Remote Sensing Digital Image Analysis: An Introduction},
  publisher = {Springer},
  year      = {2013}
}

@article{zhang2015hyperspectral,
  author  = {L. Zhang and L. Zhang and D. Tao and X. Huang},
  title   = {Hyperspectral Remote Sensing Image Classification via Multiple Feature Learning},
  journal = {IEEE Transactions on Geoscience and Remote Sensing},
  volume  = {53},
  number  = {3},
  pages   = {1592--1606},
  year    = {2015}
}

@article{liu2020dimensionality,
  title={Dimensionality reduction of hyperspectral images based on improved spatial--spectral weight manifold embedding},
  author={Liu, Hong and Xia, Kewen and Li, Tiejun and Ma, Jie and Owoola, Eunice},
  journal={Sensors},
  volume={20},
  number={16},
  pages={4413},
  year={2020},
  publisher={MDPI}
}

@article{yun2023spectr,
  title={SpecTr: Spectral transformer for microscopic hyperspectral pathology image segmentation},
  author={Yun, Boxiang and Lei, Baiying and Chen, Jieneng and Wang, Huiyu and Qiu, Song and Shen, Wei and Li, Qingli and Wang, Yan},
  journal={IEEE Transactions on Circuits and Systems for Video Technology},
  volume={34},
  number={6},
  pages={4610--4624},
  year={2023},
  publisher={IEEE}
}

@article{demir2014active,
  author  = {B. Demir and L. Bruzzone},
  title   = {A Novel Active Learning Method for Hyperspectral Image Classification},
  journal = {IEEE Transactions on Geoscience and Remote Sensing},
  volume  = {52},
  number  = {11},
  pages   = {7046--7061},
  year    = {2014}
}

@article{rope,
  author       = {Jianlin Su and
                  Yu Lu and
                  Shengfeng Pan and
                  Bo Wen and
                  Yunfeng Liu},
  title        = {RoFormer: Enhanced Transformer with Rotary Position Embedding},
  journal      = {CoRR},
  volume       = {abs/2104.09864},
  year         = {2021},
  url          = {https://arxiv.org/abs/2104.09864},
  eprinttype   = {arXiv},
  eprint       = {2104.09864},
  bibsource    = {dblp computer science bibliography, https://dblp.org}
}

@article{paviauniversity,
  author  = {D. Landgrebe},
  title   = {Hyperspectral Image Data Analysis},
  journal = {IEEE Signal Processing Magazine},
  volume  = {19},
  number  = {1},
  pages   = {17--28},
  year    = {2002},
  note    = {Pavia University dataset collected by the ROSIS-3 sensor, University of Pavia, Italy}
}

@misc{salinas,
  author = {{AVIRIS}},
  title  = {Salinas Hyperspectral Dataset},
  year   = {1998}
}

@article{hybridsn,
  author  = {A. Roy and S. Krishna and M. M. R. Krishna},
  title   = {HybridSN: Exploring 3D–2D CNN Feature Hierarchy for Hyperspectral Image Classification},
  journal = {IEEE Geoscience and Remote Sensing Letters},
  volume  = {17},
  number  = {2},
  pages   = {277--281},
  year    = {2020}
}

@article{swinhsi,
  author  = {H. Cao and Y. Wang and J. Chen and J. Wang},
  title   = {Swin-HSI: A Tiny Swin Transformer for Hyperspectral Image Classification},
  journal = {Remote Sensing},
  volume  = {14},
  number  = {13},
  pages   = {3035},
  year    = {2022}
}

@article{mobilenet,
  author  = {Andrew G. Howard and Menglong Zhu and Bo Chen and Dmitry Kalenichenko and Weijun Wang and Tobias Weyand and Marco Andreetto and Hartwig Adam},
  title   = {MobileNets: Efficient Convolutional Neural Networks for Mobile Vision Applications},
  journal = {Remote Sensing},
  volume  = {14},
  number  = {13},
  pages   = {3035},
  year    = {2022}
}

@misc{spectralmamba,
      title={SpectralMamba: Efficient Mamba for Hyperspectral Image Classification}, 
      author={Jing Yao and Danfeng Hong and Chenyu Li and Jocelyn Chanussot},
      year={2024},
      eprint={2404.08489},
      archivePrefix={arXiv},
      primaryClass={cs.CV},
      url={https://arxiv.org/abs/2404.08489}, 
}

@ARTICLE{hit,
  author={Yang, Xiaofei and Cao, Weijia and Lu, Yao and Zhou, Yicong},
  journal={IEEE Transactions on Geoscience and Remote Sensing}, 
  title={Hyperspectral Image Transformer Classification Networks}, 
  year={2022},
  volume={60},
  number={},
  pages={1-15},
  doi={10.1109/TGRS.2022.3171551}}

@article{houston,
  author={Debes, Christian and Merentitis, Andreas and Heremans, Roel and Hahn, Juergen and Frangiadakis, Nikolaos and van Kasteren, Tim and Liao, Wenzhi and Bellens, Rik and Pizurica, Aleksandra and Gautama, Sidhartha and Philips, Wilfried and Prasad, Saurabh and Du, Qian and Pacifici, Fabio},
  journal={IEEE Journal of Selected Topics in Applied Earth Observations and Remote Sensing}, 
  title={Hyperspectral and LiDAR Data Fusion: Outcome of the 2013 GRSS Data Fusion Contest}, 
  year={2014},
  volume={7},
  number={6},
  pages={2405-2418},
  doi={10.1109/JSTARS.2014.2305441}
}

@misc{indian_pines_1992,
  author = {Baumgardner, M. F. and Biehl, L. L. and Landgrebe, D. A.},
  title = {220 Band AVIRIS Hyperspectral Image Data Set: June 12, 1992 Indian Pine Test Site 3},
  year = {2015},
  publisher = {Purdue University Research Repository},
  doi = {10.4231/R7RX991C},
  howpublished = {\url{https://purr.purdue.edu/publications/1947/1}}
}

@data{houston18,
doi = {10.21227/4y56-7870},
url = {https://ieee-dataport.org/open-access/2018-ieee-grss-data-fusion-challenge-fusion-multispectral-lidar-and-hyperspectral-data},
author = {NCALM and University of Houston and IEEE GRSS},
publisher = {IEEE Dataport},
title = {2018 IEEE GRSS Data Fusion Challenge – Fusion of Multispectral LiDAR and Hyperspectral Data},
year = {2020}
}

@article{whu_hi_datasets,
title = {WHU-Hi: UAV-borne hyperspectral with high spatial resolution (H2) benchmark datasets and classifier for precise crop identification based on deep convolutional neural network with CRF},
journal = {Remote Sensing of Environment},
volume = {250},
pages = {112012},
year = {2020},
issn = {0034-4257},
doi = {https://doi.org/10.1016/j.rse.2020.112012},
url = {https://www.sciencedirect.com/science/article/pii/S0034425720303825},
author = {Yanfei Zhong and Xin Hu and Chang Luo and Xinyu Wang and Ji Zhao and Liangpei Zhang}
}

@article{ssftt,
  title={Spectral--Spatial Feature Tokenization Transformer for Hyperspectral Image Classification},
  author={Sun, Le and Zhao, Guangrui and Wu, Zebin and Zhan, Tianzeng and Liu, Wei and Luo, Jiantong},
  journal={IEEE Transactions on Geoscience and Remote Sensing},
  volume={60},
  pages={1--14},
  year={2022},
  publisher={IEEE},
  doi={10.1109/TGRS.2022.3144158}
}


\begin{IEEEbiographynophoto}{Gaurav Sharma}
Fifth-year Ph.D. candidate at the University of Arizona, USA, majoring in the College of Information Science with a minor in Statistics. Primarily, research centers on Large Vision(Video)-Language Models, Generative and Agentic AI, Natural Language Processing, Computer Vision, and Graph Neural Networks, with applications in Medical Surgical AI and Education AI.
\end{IEEEbiographynophoto}

\begin{IEEEbiographynophoto}{Eungjoo Lee}
(Member, IEEE) received a Ph.D. degree from the University of Maryland, College Park, MD, USA, in 2021. \\
Following his doctoral studies, he served as a postdoctoral research fellow at Massachusetts General Hospital/Harvard Medical School. He is currently an Assistant Professor in the Department of Electrical and Computer Engineering and Division Chief of Ophthalmic AI in the Department of Ophthalmology and Vision Science at the University of Arizona, Tucson, AZ, USA. He is also affiliated with the BIO5 Institute and the UA Cancer Center, and serves as a scientific advisor to the Brain \& Body Imaging Center. He conducts research on grounded intelligence systems that bridge perception and reliable decision-making in real-world environments. 
\end{IEEEbiographynophoto}

\vfill

\end{document}